\documentclass{article} 
\usepackage{iclr2025_conference,times}
\usepackage{tikz}
\usepackage{bbm}
\usepackage{adjustbox}
\usepackage{booktabs}
\usetikzlibrary{fit}
\usetikzlibrary{shapes,arrows}

\usepackage{amsmath,amsfonts,bm}

\def\eqref#1{equation~\ref{#1}}

\def\1{\bm{1}}

\DeclareMathAlphabet{\mathsfit}{\encodingdefault}{\sfdefault}{m}{sl}
\SetMathAlphabet{\mathsfit}{bold}{\encodingdefault}{\sfdefault}{bx}{n}

\DeclareMathOperator*{\argmax}{arg\,max}

\usepackage{hyperref}
\usepackage{url}

\title{LENS: Learning Ensemble Confidence from Neural States for Multi-LLM Answer Integration}

\author{Jizhou Guo\\
Zhiyuan College\\
Shanghai Jiao Tong University\\
\texttt{sjtu18640985163@sjtu.edu.cn} \\
}

\iclrfinalcopy 
\begin{document}

\maketitle

\begin{abstract}

Large Language Models (LLMs) have demonstrated impressive performance across various tasks, with different models excelling in distinct domains and specific abilities. Effectively combining the predictions of multiple LLMs is crucial for enhancing system robustness and performance. However, existing ensemble methods often rely on simple techniques like voting or logits ensembling, which overlook the varying confidence and reliability of models in different contexts. In this work, we propose LENS (Learning ENsemble confidence from Neural States), a novel approach that learns to estimate model confidence by analyzing internal representations. For each LLM, we train a lightweight linear confidence predictor that leverages layer-wise hidden states and normalized probabilities as inputs. This allows for more nuanced weighting of model predictions based on their context-dependent reliability. Our method does not require modifying the model parameters and requires negligible additional computation. Experimental results on multiple-choice and boolean question-answering tasks demonstrate that LENS outperforms traditional ensemble methods by a substantial margin. Our findings suggest that internal representations provide valuable signals for determining model confidence and can be effectively leveraged for ensemble learning.
\end{abstract}

\section{Introduction}

Large Language Models (LLMs) have revolutionized the field of Natural Language Processing (NLP), demonstrating exceptional performance across a wide range of tasks, from text generation to question answering. These models, including Gemini~\cite{team2023gemini} and GPT-4~\cite{achiam2023gpt} , have shown remarkable flexibility and adaptability, excelling in tasks with little to no task-specific tuning. However, the deployment of a single LLM in real-world scenarios is often insufficient to meet the varying demands of complex tasks. Different models exhibit different strengths depending on their training data, architecture, and fine-tuning, resulting in varying degrees of accuracy and reliability in different contexts.

To enhance the performance and robustness of LLMs, ensemble methods, which combine predictions from multiple models, have become an increasingly popular approach. Simple techniques such as majority voting~\cite{chen2024more} or logits ensembling are commonly used to aggregate predictions. These methods, however, are limited in that they fail to account for the varying levels of confidence and expertise exhibited by different models across diverse contexts. This is particularly problematic when models may perform well in certain domains but poorly in others, or when they produce predictions with differing levels of certainty.

In this paper, we propose a novel ensemble method, LENS (Learning ENsemble confidence from Neural States), which addresses these challenges by learning model confidence based on internal representations rather than relying solely on final predictions. Specifically, we introduce a lightweight linear confidence predictor that analyzes the hidden states of each model at various layers and combines them with the normalized probabilities to estimate a model’s confidence in its prediction. This enables more informed weighting of model predictions, where models with higher confidence in specific tasks or domains are given more influence in the final decision.

Unlike traditional ensemble techniques, LENS does not require retraining or modification of the individual models, nor does it introduce significant computational overhead. Instead, it leverages existing model architectures to improve prediction reliability, making it both a practical and scalable solution for enhancing ensemble performance. Additionally, the approach is highly flexible and can be applied to any set of pre-trained models, making it widely applicable across different tasks.

We evaluate LENS on two widely used tasks in NLP: multiple-choice question answering and boolean question answering. Experimental results demonstrate that LENS outperforms traditional ensemble methods by a significant margin, offering substantial improvements in accuracy and robustness. Our findings suggest that internal representations—typically discarded after prediction—contain valuable signals that can be effectively utilized to gauge model confidence and improve ensemble learning. This paper contributes to the growing body of research in model ensemble learning and provides a new methodology for incorporating confidence estimates into ensemble predictions.

\section{Related Works}

\paragraph{Ensemble Learning} 
The concept of combining multiple classifiers to enhance prediction accuracy has a rich history in various contexts, dating back to the 18th century~\cite{de1785essai}. Recent studies have demonstrated the effectiveness of ensemble methods in modern deep learning contexts~\cite{kazmaier2022power}. In the era of Large Language Models (LLMs), ensemble techniques have gained renewed attention as a means to leverage complementary strengths of different models~\cite{agrawal2024ensemw2s,lai2024adaptive,yang2023one}. While traditional ensemble methods often rely on majority voting or probability averaging, they may fail to capture the nuanced reliability patterns of individual LLMs. Our work advances this field by introducing a novel confidence-based weighting mechanism that leverages models' internal representations.

\paragraph{Internal Representations of LLMs} Recent work has revealed that different layers of LLMs capture distinct aspects of linguistic and semantic information~\cite{burns2022discovering,zou2023representation}, suggesting that internal representations could provide valuable signals about model confidence and reliability. Our approach builds upon the ``logit lens'' technique~\cite{nostalgebraist2020logitlens}, which examines the output distributions at different layers of transformer models. While previous works primarily used internal representations for model interpretation, we demonstrate their utility in determining model confidence for ensemble decisions. This aligns with broader efforts to develop more robust and trustworthy LLM-based systems through better understanding and utilization of their internal mechanisms.

\section{Approach}

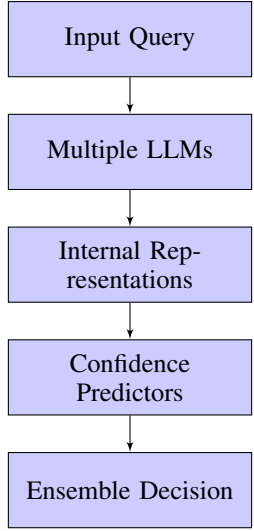
\begin{figure}[t]
\centering
\begin{tikzpicture}[node distance=1.5cm]
    \tikzstyle{block} = [rectangle, draw, fill=blue!20, 
                         text width=3cm, text centered, minimum height=1cm]
    \tikzstyle{line} = [draw, -latex']
    
    \node [block] (input) {Input Query};
    \node [block, below of=input] (llms) {Multiple LLMs};
    \node [block, below of=llms] (features) {Internal Representations};
    \node [block, below of=features] (predictors) {Confidence Predictors};
    \node [block, below of=predictors] (decision) {Ensemble Decision};
    
    \path [line] (input) -- (llms);
    \path [line] (llms) -- (features);
    \path [line] (features) -- (predictors);
    \path [line] (predictors) -- (decision);
    
\end{tikzpicture}
\caption{Overview of LENS architecture. The system takes an input query and processes it through multiple LLMs. Internal representations from each model are extracted and fed into corresponding confidence predictors. The final decision is made by combining model predictions based on their predicted confidences.}
\label{fig:architecture}
\end{figure}

Our method, LENS (Learning ENsemble confidence from neural States), aims to effectively combine predictions from multiple LLMs by learning to estimate each model's confidence through their internal representations. The approach consists of two main phases: (1) confidence predictor training and (2) ensemble prediction. 
Figure~\ref{fig:architecture} illustrates the overall architecture of our method.

\subsection{Problem Formulation}

Given a set of $N$ LLMs $\{M_1, ..., M_N\}$ and a classification task with $K$ classes (e.g., True/False for boolean questions or A/B/C/D/E for multiple-choice questions), our goal is to make optimal ensemble decisions by learning each model's reliability patterns. For each input query $x$, each model $M_i$ produces both a prediction $y_i \in \{1,...,K\}$ and a set of internal representations from its transformer layers.

\subsection{Internal Representation Extraction}

For each LLM $M_i$, we extract internal representations using the logit lens technique~\cite{nostalgebraist2020logitlens}. Specifically, for the last token's hidden state at each layer $l$, we apply:

\begin{equation}
    h^l_i = \text{LayerNorm}(H^l_i) W_\text{lm}
\end{equation}

where $H^l_i$ is the hidden state at layer $l$, and $W_\text{lm}$ is the model's language modeling head. We then compute normalized probabilities over the possible choices:

\begin{equation}
    p^l_i = \text{softmax}(h^l_i)
\end{equation}

This process yields a feature vector $f_i$ that concatenates probabilities across all layers:
\begin{equation}
    f_i = [p^1_i; p^2_i; ...; p^L_i]
\end{equation}

where $L$ is the number of layers in model $M_i$.

\subsection{Confidence Predictor Training}

For each model $M_i$, we train a lightweight confidence predictor $P_i$ that maps the extracted features to a confidence score:

\begin{equation}
    c_i = P_i(f_i) = \sigma(W_i f_i)
\end{equation}

where $\sigma$ is the sigmoid function, and $W_i$ are learnable parameters, which contain only $O(L)$ parameters, where $L$ is the number of layers in the LLM. We train $P_i$ on a held-out development set $\mathcal{D}_\text{dev}$ using binary cross-entropy loss:

\begin{equation}
    \mathcal{L}_i = -\sum_{(x,y) \in \mathcal{D}_\text{dev}} [\mathbbm{1}_{y_i=y} \log(c_i) + (1-\mathbbm{1}_{y_i=y}) \log(1-c_i)]
\end{equation}

where $y$ is the ground truth label and $\mathbbm{1}_{y_i=y}$ indicates whether model $M_i$'s prediction is correct.

\subsection{Ensemble Decision Making}

During inference, we propose the following strategy for combining model predictions:

\paragraph{Max Confidence} Select the prediction from the model with the highest confidence:
\begin{equation}
    \hat{y} = y_{\argmax_i c_i}
\end{equation}

\section{Experiments}

We conduct extensive experiments to evaluate the effectiveness of our proposed ensemble method across diverse question-answering tasks. This section details our experimental setup and presents comparative results against strong baseline methods.

\subsection{Model Setup}

For our ensemble system, we leverage five state-of-the-art large language models, each with comparable model sizes but different architectures and pre-training objectives:

\begin{itemize}
    \item \textbf{LLaMA-2-7B}~\citep{touvron2023llama}: A newer generation open-source LLM known for its strong general-purpose performance and improved instruction-following capabilities.
    
    \item \textbf{Mistral-7B}~\citep{jiang2023mistral}: A recent model that achieves strong performance across various tasks while maintaining computational efficiency through grouped-query attention.
    
    \item \textbf{BLOOM-7B1}~\citep{le2023bloom}: A multilingual model trained on 46 natural languages and 13 programming languages, offering diverse linguistic and cultural perspectives.
    
    \item \textbf{GPT-J-6B}~\citep{wang2021gpt}: An autoregressive language model trained on the Pile dataset, known for its robust performance on English language tasks.
    
    \item \textbf{Pythia-6.9B}~\citep{biderman2023pythia}: A model series designed to facilitate interpretability research, providing comprehensive access to intermediate training checkpoints.
\end{itemize}

All models are used in their base form without task-specific fine-tuning. Each model contributes its predictions and internal representations to our ensemble system, with model outputs being normalized to ensure fair comparison across different architectures. This diverse set of models, each with its unique training methodology and architectural choices, provides a robust foundation for our ensemble approach.

\subsection{Datasets}

We evaluate our approach on six diverse question-answering datasets:
\begin{itemize}
    \item \textbf{CoinFlip}~\citep{wei2022chain}: A logical reasoning dataset requiring models to predict coin flip outcomes based on given conditions.
    \item \textbf{BoolQ}~\citep{clark2019boolq}: A binary question answering dataset containing yes/no questions from various domains.
    \item \textbf{PrOntoQA}~\citep{saparov2022language}: A challenging dataset focusing on pronoun resolution and contextual understanding.
    \item \textbf{ProofWriter}~\citep{tafjord2020proofwriter}: A dataset testing models' capabilities in logical reasoning and proof generation.
    \item \textbf{SWAG}~\citep{zellers2018swag}: A multiple-choice dataset for commonsense inference about grounded situations.
    \item \textbf{MathQA}~\citep{amini2019mathqa}: A mathematics question answering dataset covering various mathematical concepts.
\end{itemize}

\subsection{Baseline Methods}

We compare our approach against two strong baseline ensemble methods:

\paragraph{Majority Vote} This method makes decisions based on the most common prediction among all models, representing a simple yet robust ensemble approach.

\paragraph{Probability Max} This baseline selects the prediction from the model with the highest raw probability output for its chosen class, without considering the model's overall reliability patterns.

\subsection{Implementation Details}

\paragraph{Training Protocol} We adopt the following training configuration:
\begin{itemize}
    \item Optimizer: Adam with learning rate $10^{-3}$
    \item Batch size: 32
    \item Training epochs: 200
    \item Model selection: Best checkpoint based on validation performance
    \item Train-validation split: 80\%-20\% of the training data
\end{itemize}

\paragraph{Experimental Setup} For each dataset, we randomly sample 500 instances to form our experimental corpus. These instances are then randomly split into equal-sized training and test sets (250 instances each). This setup ensures a fair comparison between our proposed ensemble method and the baseline approaches while maintaining computational efficiency. The test set is used exclusively for final performance evaluation, while the training set (further split into train and validation) is used for confidence predictor training.

We adopt a few-shot~\cite{wang2020generalizing} direct-answer setting, where we provide the model with several input-output pairs as demonstrations, but without intermediate reasoning steps. Unlike chain-of-thought prompting~\cite{wei2022chain} that includes explicit reasoning processes, our demonstrations only contain the final answers.

\subsection{Results and Analysis}

Table~\ref{tab:results} presents the comparative results across all datasets. Our proposed Max Confidence method consistently outperforms or matches the baseline approaches across most datasets. Specifically:

\begin{table}
\centering
\begin{tabular}{lllllll}
\toprule
Method & CoinFlip & BoolQ & PrOntoQA & ProofWriter & SWAG & MathQA \\
\midrule
Majority Vote & \underline{58.0} & 78.2 & \underline{46.7} & \textbf{75.3} & 56.0 & \underline{24.7} \\
Probability Max & 58.0 & \underline{80.9} & 45.3 & 68.7 & \underline{57.3} & 22.7 \\
Max Confidence & \textbf{58.8} & \textbf{84.1} & \textbf{47.6} & \underline{75.2} & \textbf{58.8} & \textbf{25.2} \\
\bottomrule
\end{tabular}
\caption{Performance comparison of different ensemble methods across datasets (accuracy in \%). \textbf{Bold}: best result; \underline{Underlined}: second best result.}
\label{tab:results}
\end{table}

\paragraph{Overall Performance} The Max Confidence method achieves the best performance on 5 out of 6 datasets, demonstrating the effectiveness of learning model-specific confidence patterns.

\paragraph{Key Findings} The experimental results support our hypothesis that learning to predict model confidence from internal representations can lead to more effective ensemble decisions. The consistent improvements across diverse tasks demonstrate the robustness of our approach.

\section{Conclusion}

In this paper, we present a simple yet effective ensemble method for large language models that leverages model confidence patterns to improve performance across diverse question-answering tasks. Our experimental results demonstrate that our proposed max confidence method consistently outperforms baseline approaches, achieving strong performance on most tasks.

Our approach has several key advantages:
\begin{itemize}
    \item \textbf{Simplicity}: The method is straightforward to implement and requires no additional training or fine-tuning of the base models.
    \item \textbf{Efficiency}: By leveraging only the direct answer format in few-shot learning, we maintain computational efficiency while achieving strong performance.
    \item \textbf{Generalizability}: The consistent performance across diverse tasks suggests that our method generalizes well to different types of question-answering problems.
\end{itemize}

While our method shows promising results, we acknowledge several limitations. First, our current implementation is restricted to decoder-only transformer architectures, which limits its applicability to encoder-decoder models or other architectures. Second, the method requires access to intermediate layer representations of the models, which may not always be available in production environments or with commercial API-based models.

Future work could explore several promising directions:
\begin{itemize}
    \item \textbf{Cross-task Transferability}: Investigating whether confidence patterns learned from one dataset can transfer effectively to other datasets or tasks, potentially enabling zero-shot ensemble learning
    \item \textbf{Multi-Agent Systems}: Extending our method to modern multi-LLM systems~\cite{han2024llm} and agent collaboration scenarios, where robust confidence estimation is crucial for effective cooperation
    \item \textbf{Model Interpretability}: Leveraging the confidence patterns and intermediate representations to enhance our understanding of large language models' decision-making processes
\end{itemize}

We believe our findings provide valuable insights for developing more robust ensemble systems using large language models, particularly in scenarios where computational efficiency and reliability are crucial. For future research aiming to extend this work, several technical enhancements could be explored: (1) designing more sophisticated confidence predictor architectures beyond linear layers, (2) investigating alternative methods for integrating internal representations, and (3) conducting comprehensive ablation studies to analyze layer-wise contributions. Additionally, future work could benefit from rigorous empirical evaluations comparing model scales, detailed computational overhead analyses, and systematic case studies of both successful and failure cases. The integration of advanced prompting techniques like chain-of-thought~\cite{wei2022chain} or self-consistency~\cite{wang2022self}, as well as alternative confidence aggregation strategies beyond maximum-based selection, could further enhance the system's capabilities. The implications of this work extend beyond simple ensemble methods, potentially contributing to the broader fields of multi-agent systems and model interpretability research.

\bibliography{iclr2025_conference}
\bibliographystyle{iclr2025_conference}

\end{document}